# New version of Gram-Schmidt Process with inverse for Signal and Image Processing

**Mario Mastriani**

mmastri@gmail.com

*Abstract*—The Gram-Schmidt Process (GSP) is used to convert a non-orthogonal basis (a set of linearly independent vectors, matrices, etc) into an orthonormal basis (a set of orthogonal, unit-length vectors, bi or tri dimensional matrices). The process consists of taking each array and then subtracting the projections in common with the previous arrays. This paper introduces an enhanced version of the Gram-Schmidt Process (EGSP) with inverse, which is useful for Digital Signal and Image Processing, among others applications.

# 1 Introduction

Orthonormalisation processes play a key role in many iterative methods used in Correlation Matrix Memory [1], Array Signal Processing [2], the Kalman Filtering problem [3], Datamining and Bioinformatics [4], among others [5, 6], with different implementation possibilities [3, 7-9], however, the issues involved in the use of Very-Large-Scale Integration (VLSI) technology to implement an adaptive version of the GSP, based on the escalator structure, are discussed in [10], while alternative versions for optimal filtering and control problems without using GSP are discussed in [11-15]. Besides, it has a very important field of applications in communications, see [16-25]. Returning to the situation at hand, the EGSP is useful to perform a stable orthonormalization with inverse process, in opposition to previous versions that achieve it thanks to impractical or unstable algorithmic methods [26, 27], or being stable doesn't have inverse, such as Modified GSP [28]. Finally, a good orthonormalisation algorithm with inverse is essential for different applications such as filtering and compression in Digital Signal and Image Processing.

The original Gram-Schmidt Process (GSP) is outlined in Section II. Inverse of GSP in Section III. Enhanced Gram-Schmidt Process (EGSP) in Section IV. Inverse of EGSP (IEGSP) in Section V. Bidimensional EGSP (EGSP2D) in Section VI. Inverse of EGSP2D (IEGSP2D) in Section VII. Performance proof is outline in Section VIII. Finally, Section IX provides a conclusion of the paper.

# 2 Gram-Schmidt Process (GSP)

First, we will explain the algorithms with sets of linearly independent vectors. However, then we extend these concepts to the case of sets of linearly independent matrices with two or three dimensions, in both cases, i.e., in algebraic and algorithmic version.

## 2.1 Algebraic version

Given a set of key vectors that are linearly independent but non-orthonormal, it is possible to use a preprocessor to transform them into an orthonormal set; the processor is designed to perform a Gram-Schmidt orthogonalisation on the key vectors prior to association [29, 30]. This way to transform is linear, maintaining a one-to-one correspondence between the input (key) vectors $v_1, v_2, \ldots, v_N$, and the resulting orthonormal vectors $u_1, u_2, \ldots, u_N$, as indicated in:

$$\{v_1, v_2, \ldots, v_N\} \Leftrightarrow \{u_1, u_2, \ldots, u_N\}$$

where $u_1 = v_1$, and the remaining $u_n$ are defined by [30]

$$u_n = v_n - \sum_{m=1}^{n-1} r_{n,m} u_m \qquad n = 2, 3, \ldots, N \qquad (1)$$

with

$$r_{n,m} = \frac{u_m^T v_n}{u_m^T u_m} \qquad (2)$$

where $(.)^T$ means transpose of $(.)$. The orthogonality of key vectors may also be approached using statistical considerations. Specifically, if the input space dimension $M$ is large and the key vectors have statistically independent elements, then they will be close to orthogonality with respect to each other [1]. However, the number (quantity) of coefficients $r_{m,n}$ is

$$N_r = \frac{N(N-1)}{2} \qquad (3)$$

which is independent of the key vectors dimension $M$, i.e., $v_i \in \mathbb{R}^M$ and $u_i \in \mathbb{R}^M$, being $N$ the number of vectors. As an example, Eq.(4) shows us the detail of matrix $r$, e.g., of 6 columns (vectors).

$$r = \begin{bmatrix} 1 & 0 & 0 & 0 & 0 & 0 \\ r_{2,1} & 1 & 0 & 0 & 0 & 0 \\ r_{3,1} & r_{3,2} & 1 & 0 & 0 & 0 \\ r_{4,1} & r_{4,2} & r_{4,3} & 1 & 0 & 0 \\ r_{5,1} & r_{5,2} & r_{5,3} & r_{5,4} & 1 & 0 \\ r_{6,1} & r_{6,2} & r_{6,3} & r_{6,4} & r_{6,5} & 1 \end{bmatrix} \quad (4)$$

In a lossy compression context, it is extremely important run a sorted pruning those components of lower weight in the reconstruction (or decompression), in this case, the rows with higher subscript. So $r$ is a matrix (in fact, square and upper triangular).

**2.2 Algorithmic version**

The algorithmic version of GSP is based on Eq. (1) and (2), as shown in Fig. 1. A matrix **v** ($M$ x $N$) would be built from the base of the input (key) vectors $v_1, v_2, \ldots, v_N$, and its columns would be the same vectors. Similarly, with the resulting orthonormal vectors $u_1, u_2, \ldots, u_N$, a matrix **u** ($M$ x $N$) is built whose columns are these vectors. The process will also build a matrix **r** ($N$ x $N$), in terms of Equations (1) to (4). The algorithm as a function in MATLAB® code is

```
function [u,r] = gsp(v)
[M,N] = size(v);
u(:,1) = v(:,1);
r = eye(N,N); % eye(•) is a built-in function of MATLAB®
for n = 2:N   % eye(•) represents identity matrix
  acu = 0;
  for m = 1:n-1
    num = dot(v(:,n),u(:,m)); % dot(•,•) represents inner product of vectors in MATLAB®
    den = dot(u(:,m),u(:,m)); % dot(•,•) is a built-in function of MATLAB®
    r(n,m) = num/den;
    acu = acu + r(n,m)*u(:,m);
  end
  u(:,n) = v(:,n) - acu;
end
```

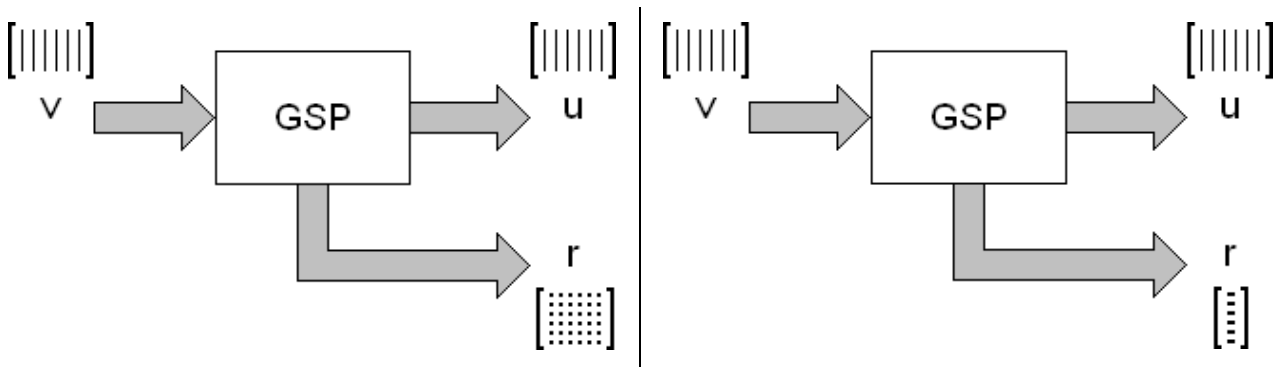

**Fig.1** Gram-Schmidt Process (GSP) with matricial (left) and vectorial (right) **r**.

On the other hand, and for a more efficient implementation of the algorithm in terms of **r**, we can implement the process by building a vector **r** ($N_r$ x 1), where, the number of coefficients (and considerations) **r** is the same of Eq.(3). However, and in a lossy compression context, the sorted pruning those components of vector

**r** may be carry out in terms of a function named: *generating function blocks r* (gfbr), whose MATLAB® code is:

```
function y = gfbr(x)
aux = 0:1:x;
y = sum(aux); % sum(•) is a built-in function of MATLAB®
```

where, *x* is the number of considered vector (column), and *y* is the vector number of coefficients considered.

Finally, the algorithm as a function in MATLAB® code according to this approach is

```
function [u,r] = gsp(v)
[M,N] = size(v);
u(:,1) = v(:,1);
r = [];
k = 0;
for n = 2:N
  acu = 0;
  for m = 1:n-1
    num = dot(v(:,n),u(:,m));
    den = dot(u(:,m),u(:,m));
    k = k+1;
    r(k) = num/den;
    acu = acu + r(k)*u(:,m);
  end
  u(:,n) = v(:,n) - acu;
end
```

## 3 Inverse of GSP (IGSP)

An algorithmic version of the IGSP is indispensable for multiple applications [1-31], therefore, an original process is exposed below. However, it is important to mention that this version is unstable under certain conditions [28].

### 3.1 Algebraic version for the discrete IGSP

Based on prior considerations $v_1 = u_1$, and considering Eq. (2), the remaining $v_n$ are defined by

$$v_n = u_n + \sum_{m=1}^{n-1} r_{n,m} u_m \quad n = 2, 3, \ldots, N \tag{5}$$

### 3.2 Algorithmic version

The algorithmic version of IGSP is based on Eq. (5) and (2), as shown in Fig. 2.

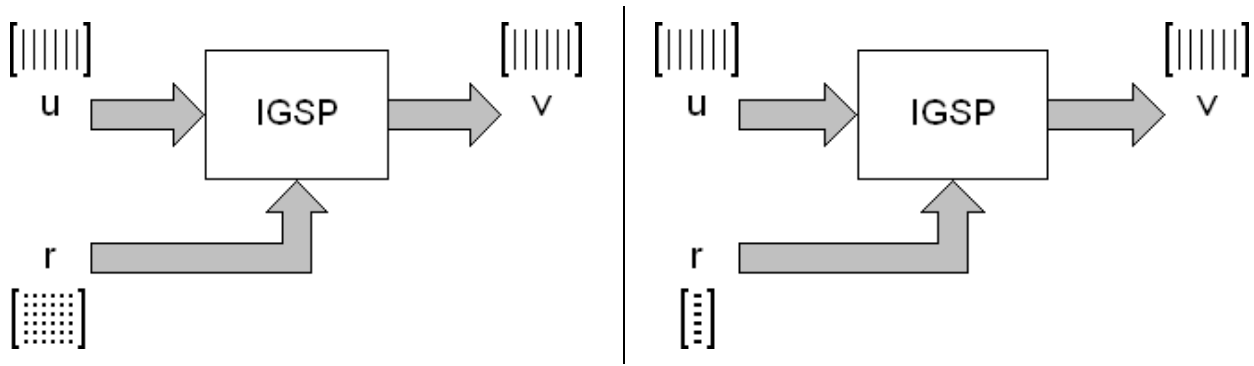

**Fig. 2** Inverse Gram-Schmidt Process (IGSP) with matricial (left) and vectorial (right) **r**.

The algorithm as a MATLAB® function (with an exclusively vector version of **r**) would be:

```
function v = igsp(u,r)
v(:,1) = u(:,1);
k = 0;
[M,N] = size(u);
for n = 2:N
  acu = 0;
  for m = 1:n-1
    k = k+1;
    acu = acu + r(k)*u(:,m);
  end
  v(:,n) = u(:,n) + acu;
end
```

Finally, a modified version of the traditional GSP exists [28] (based on QR decomposition), and it is stable, but it doesn't have inverse because it is an in-situ method, i.e., it is a destructive algorithm [31]. This is the reason for which becomes indispensable a new algorithm that being stable has inverse.

### 4 Enhanced Gram-Schmidt Process (EGSP)

The mentioned algorithm is the very traditional version of GSP (well known for its bad numerical properties, see [28]), modified versions of the same one exist, called Modified GSP (MGSP), see [31], but unfortunately, they don't have inverse, because, they are in-situ algorithm, i.e., they are applicable on the same set of key vectors, i.e., they constitute destructive methods. On the other hand, the unstability of GSP happens when certain conditions occur, and they are relating to numerical analysis [31].

### 4.1 Algebraic version of EGSP

Therefore, and to assure the orthogonality it is necessary to apply the following procedure on the input (key) vectors $v_1, v_2, \ldots, v_N$, with $u_1 = v_1$:

$$u_n = u_n - \sum_{m=1}^{n-1} r_{n,m} u_m \quad n = 1, 2, \ldots, N \tag{6}$$

$$r_{n,m} = \frac{v_n^T u_m}{u_m^T u_m} \tag{7}$$

Although the change in these equations seems insignificant, the results are comparatively impressive.

### 4.2 Algorithmic version of EGSP

The algorithm as a MATLAB® function would be:

```
function [u,r] = egsp(v)
[M,N] = size(v);
r = [];
k = 0;
for n = 1:N
  u(:,n) = v(:,n);
  for m = 1:n-1
    num = dot(v(:,n),u(:,m));
    den = dot(u(:,m),u(:,m));
    k = k+1;
    r(k) = num/den;
    u(:,n) = u(:,n)-r(k)*u(:,m);
  end
end
```

## 5 Inverse of EGSP (IEGSP)

With similar considerations regarding to IGSP, we present IEGSP, in its algebraic and algorithmic forms.

### 5.1 Algebraic version of IEGSP

Based on prior considerations $v_1 = u_1$, and considering Eq. (2), the remaining $v_n$ are defined by

$$u_n = u_n + \sum_{m=1}^{n-1} r_{n,m} u_m \quad n = 1, 2, \ldots, N \tag{8}$$

### 5.2 Algorithmic version of IEGSP

The algorithm as a MATLAB® function would be:

```
function v = iegsp(u,r)
[M,N] = size(u);
k = 0;
for n = 1:N
  v(:,n) = u(:,n);
  for m = 1:n-1
    k = k+1;
    v(:,n) = v(:,n)+r(k)*u(:,m);
  end
end
```

## 6 Two-dimensional EGSP (EGSP2D)

Making similar considerations to the vector case, we can apply a two-dimensional version of EGSP to two-dimensional blocks too, and linearly independents, with a configuration like to Fig.3.

Here too, the number (quantity) of coefficients $r$ is independent of the key blocks dimension $M{\times}B$, i.e., $v_i \in \mathbb{R}^{M \times B}$ and $u_i \in \mathbb{R}^{M \times B}$, being $N$ the number of blocks. Besides, such quantity is the same to Eq.(3).

The algorithm as a MATLAB® function (with an exclusively vector version of **r**) would be a natural extension of the vector version, and indeed it is.

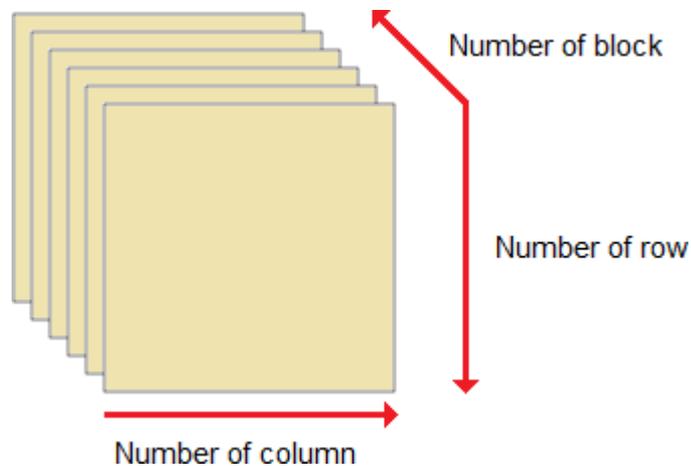

**Fig. 3** Three-dimensional matrix **v**.

```
function [u,r] = egsp2d(v)
[M,B,N] = size(v);
k = 0;
r = [];
for n = 1:N
  u(:,:,n) = v(:,:,n);
  for m = 1:n-1
    num = sum(dot(v(:,:,n),u(:,:,m))); % sum(dot(•,•)) represents inner product of
    den = sum(dot(u(:,:,m),u(:,:,m))); % matrices in MATLAB®
    k = k+1;
    r(k) = num/den;
    u(:,:,n) = u(:,:,n)-r(k)*u(:,:,m);
  end
end
```

Identical considerations can be made for the case of three, four and more dimensions.

## 7 Inverse of EGSP2D (IEGSP2D)

The algorithm as a MATLAB® function would be:

```
function v = icgs2d(u,r)
[M,B,N] = size(u);
k = 0;
for n = 1:N
  v(:,:,n) = u(:,:,n);
  for m = 1:n-1
    k = k+1;
    v(:,:,n) = v(:,:,n)+r(k)*u(:,:,m);
  end
end
```

## 8 Performance proof

In this work, the assessment parameters that are used to evaluate the performance of speckle reduction are: Mean Absolute Error (MAE), Mean Squared Error (MSE), Peak Signal-To-Noise Ratio (PSNR), and Proof of Orthogonality (PO). The first three test the reverse attribute, while the latter checks the quality of the orthogonalisation process.

*Mean Absolute Error (MAE)*
The mean absolute error is a quantity used to measure how close forecasts or predictions are to the eventual outcomes. The mean absolute error (MAE) is given by

$$MAE = \frac{1}{MxN} \sum_{m=0}^{M-1} \sum_{n=0}^{N-1} \|v(m,n)-\hat{v}(m,n)\| \tag{9}$$

which for two $NR \times NC$ (rows-by-columns) group of column vectors $v$ and $\hat{v}$, where the second one of the images is considered a decompressed approximation of the other of the first one.

*Mean Squared Error (MSE)*
The mean square error or MSE in Signal and Image Compression is one of many ways to quantify the difference between an original array and the true value of the quantity being decompressed array, which for two $M \times N$ (rows-by-columns) group of column vectors $v$ and $\hat{v}$, where the second one is considered a recovered approximation of the other is defined as:

$$MSE = \frac{1}{MxN} \sum_{m=0}^{M-1} \sum_{n=0}^{N-1} \|v(m,n)-\hat{v}(m,n)\|^2 \tag{10}$$

*Peak Signal-To-Noise Ratio (PSNR)*

The phrase peak signal-to-noise ratio, often abbreviated PSNR, is an engineering term for the ratio between the maximum possible power of a signal and the power of corrupting noise that affects the fidelity of its representation. Because many signals have a very wide dynamic range, PSNR is usually expressed in terms of the logarithmic decibel scale. The PSNR is most commonly used as a measure of quality of reconstruction in image compression, etc [20]. It is most easily defined via the mean squared error (MSE), so, the PSNR is defined as [20]:

$$PSNR = 10\,log_{10}\left(\frac{MAX_v^2}{MSE}\right) = 20\,log_{10}\left(\frac{MAX_v}{\sqrt{MSE}}\right) \tag{11}$$

Here, $MAX_v$ is the maximum value of $v$. In Digital Image Processing, and when the pixels are represented using 8 bits per sample, this is 256. More generally, when samples are represented using linear pulse code modulation (PCM) with B bits per sample, maximum possible value of $MAX_I$ is $2^B$-1.

For color images with three red-green-blue (RGB) values per pixel, the definition of PSNR is the same except the MSE is the sum over all squared value differences divided by image size and by three [20].

Typical values for the PSNR in lossy image and video compression are between 30 and 50 dB, where higher is better.

*Proof of orthogonality (PO)*

As we have mentioned before, this metric checks the quality of the orthogonalisation process. On the other hand, it is a vector, and each and every one of its elements must be virtually zero for the purposes of verifying and indirectly also the stability of the algorithm, especially as *N* (number of vectors or blocks) grows.

$$PO_k = u_n^T u_m \,,\, \forall k \in \left[1, \frac{(N-1)N}{2}\right],\, \wedge\, \forall n \neq m\, /\, n \wedge m \in [1, N] \tag{12}$$

That is, it is a vector of inner products of all against all except themselves. In fact, this vector has a length equal to $N(N-1)/2$ pairs of components from group of orthogonalized column vectors **u**.

The routine of Proof of Orthogonality (PO) in MATLAB® is explained below, in fact, it is the last of the three routines that are detailed below.

On the other hand, the used routines are the functions: egsp() and iegsp(), besides the following ones:

```
% main file
M = input('M = ');
N = input('N = ');
v = rand(M,N); % rand(•,•) is a built-in function of MATLAB®
               % rand(•,•) represents a pseudorandom scalar drawn from the
               % standard uniform distribution on the open interval (0,1)
[u,r] = egsp(v);
w = po(u);     % proof of orthogonality
max_po_w = max(abs(w)) % max(•) is a built-in function of MATLAB®
                       % max(•) returns the largest elements along different
                       % dimensions of an array.
plot(1:length(w),w,'r*')
hold on
plot(1:length(w),w,'b')
title('Proof of orthogonality')
axis([ 1 length(w) min(w) max(w) ])
ve = iegsp(u,r);
[mae,mse,psnr] = metrics(v,ve)
```

```
function [mae,mse,psnr] = metrics(x,y)
Rx = length(x); % length(•) is a built-in function of MATLAB®
Ry = length(y); % length(•) finds the number of elements along the largest
                % dimension of an array.
if Rx ~= Ry,
  error('Los vectores tienen diferentes dimensiones');
end
mae = sum(sum(abs(x-y)))/Rx; % abs(•) is a built-in function of MATLAB®
mse = sum(sum((x-y).^2))/Rx; % abs(•) returns an array such that each element
                             % of such array is the absolute value of the
                             % corresponding element of (•).
if(mse ~= 0)
  psnr = 10*log10(max(max(abs(x)))^2/mse); % log10(•) is a built-in function of MATLAB®
                                           % log10(•) returns the base 10 logarithm
                                           % of the elements of X.
else
  psnr = Inf; % Inf means infinity
end

function z = po(u)
k = 0;
z = [];
[M,N] = size(u); % size(•) is a built-in function of MATLAB®
                 % size(•) returns the sizes of each dimension of array (•).
for n = 1:N-1
  for m = n+1:N
    k = k+1;
    z(k) = dot(u(:,n),u(:,m));
  end
end
```

The first routine is the main file in MATLAB® code. The second one represents the three metrics relatives to test of reversibility. Finally, the last one represents the Proof of Orthogonality.

TABLE I
PROOF OF ORTHOGONALITY VS NUMBER OF VECTORS

| Number of vectors (N) | Proof of orthogonality (maximum value) |
|---|---|
| 5 | 5.2736e-16 |
| 10 | 1.2872e-15 |
| 15 | 2.0331e-15 |
| 20 | 3.4452e-15 |

TABLE II
INVERSE TEST VIA MAE/MSE/PSNR VS NUMBER OF VECTORS.

| Number of vectors (N) | MAE | MSE | PSNR |
|---|---|---|---|
| 5 | 5.2996e-17 | 5.2772e-33 | 322.7036 |
| 10 | 3.5510e-16 | 3.5813e-32 | 314.4537 |
| 15 | 6.3300e-16 | 7.6830e-32 | 311.0503 |
| 20 | 1.1345e-15 | 1.6361e-31 | 307.8602 |

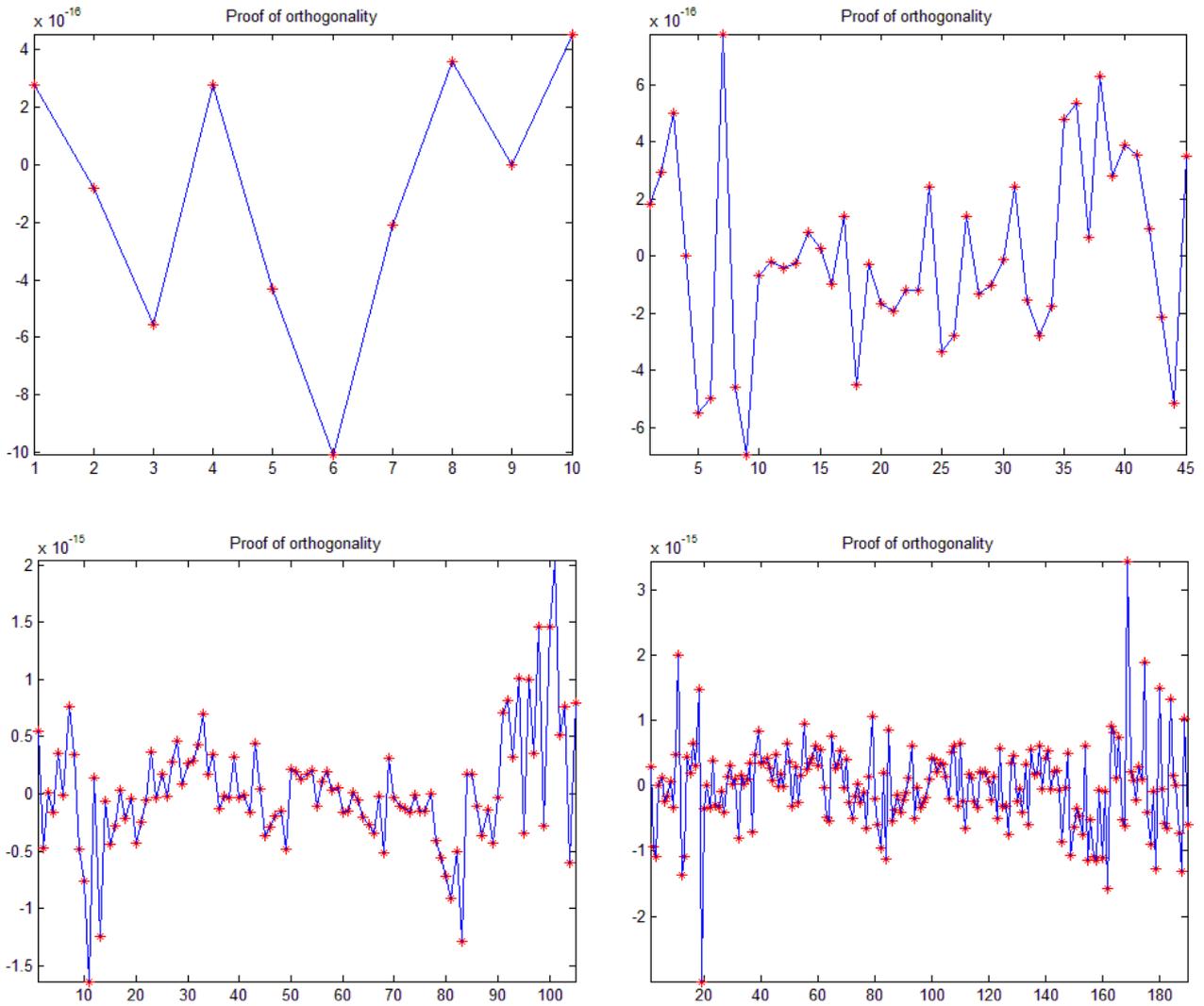

**Fig. 4** Proof of orthonormality via function po(•) for M = 20, and N = 5 (top-left), N = 10 (top-right), N = 15 (bottom-left), and N = 20 (bottom-right).

In all tests we use vectors with dimension M = 20, where, Table I shows the maximum value of the Proof of Orthogonality vector in terms of the number of vectors, while, Table II shows inverse test via MAE, MSE, and PSNR, in terms of the number of vectors, too. Finally, Fig.4 shows the Proof of Orthogonality vector in terms of the number of couples (one by one). As we can see, in all cases the results are exceptionally good.

## 9 Conclusions

In this paper, we introduce a new version of the traditional Gram Schmidt Process of orthogonalisation. However, this novel version is computationally parallelizable (making it ideal for deployment in GPGPU), but also has inverse, and is extremely fast.

All simulations (without exception) show the quality and performance of the new version, as orthogonalize and for its inverse attribute, which makes it ideal candidate for application in signal and image filtering and compression.

Finally, the classical simulations were implemented in MATLAB® R2015a (Mathworks, Natick, MA) [49] on a notebook with Intel® Core(TM) i7-4702 MQ CPU 2.20 GHz and 8 GB RAM on Microsoft® Windows 7© Ultimate 64 bits.